\documentclass[letterpaper, 10 pt, conference]{ieeeconf}  

\IEEEoverridecommandlockouts                              

\usepackage{cite}
\usepackage{amsmath,amssymb,amsfonts}
\usepackage{algorithmic}
\usepackage{algorithm}

\usepackage{graphicx}
\usepackage{textcomp}
\usepackage{xcolor}
\usepackage{verbatim}
\usepackage{multirow}
\usepackage{booktabs}
\usepackage{makecell}
\usepackage{hyperref}

\usepackage[table]{xcolor}
\usepackage[table,dvipsnames]{xcolor}
\usepackage{tikz}  
\usetikzlibrary{arrows.meta,positioning,shapes.geometric}

\usepackage{siunitx}
\definecolor{mygreen}{RGB}{218,235,192}

\title{\LARGE \bf
StepNav: Structured Trajectory Priors for Efficient and Multimodal Visual Navigation
}

\author{Xubo Luo$^{1}$, Aodi Wu$^{1}$, Haodong Han$^{1}$, Xue Wan$^{2*}$, Wei Zhang$^{2}$, Leizheng Shu$^{2}$ and Ruisuo Wang$^{2}$
\thanks{This work was supported by the National Natural Science Foundation of China (Grant No. 42171445).}
\thanks{$^{1}$Xubo Luo, Aodi Wu, and Haodong Han are with the University of Chinese Academy of Sciences, Beijing 100101, China
        {\tt\small (luoxubo23, wuaodi20, hanhaodong23)@mails.ucas.ac.cn}}%
\thanks{$^{2}$Xue Wan, Wei Zhang, Leizheng Shu, and Ruisuo Wang are with the Technology and Engineering Center for Space Utilization, Chinese Academy of Sciences, Beijing 100094, China
        {\tt\small (wanxue, zhangwei, shuleizheng wangruisuo)@cas.ac.cn}}%
}

\begin{document}

\maketitle
\thispagestyle{empty}
\pagestyle{empty}

\begin{abstract}
Visual navigation is fundamental to autonomous systems, yet generating reliable trajectories in cluttered and uncertain environments remains a core challenge. Recent generative models promise end-to-end synthesis, but their reliance on unstructured noise priors often yields unsafe, inefficient, or unimodal plans that cannot meet real-time requirements. We propose StepNav, a novel framework that bridges this gap by introducing structured, multimodal trajectory priors derived from variational principles. StepNav first learns a geometry-aware success probability field to identify all feasible navigation corridors. These corridors are then used to construct an explicit, multi-modal mixture prior that initializes a conditional flow-matching process. This refinement is formulated as an optimal control problem with explicit smoothness and safety regularization. By replacing unstructured noise with physically-grounded candidates, StepNav generates safer and more efficient plans in significantly fewer steps. Experiments in both simulation and real-world benchmarks demonstrate consistent improvements in robustness, efficiency, and safety over state-of-the-art generative planners, advancing reliable trajectory generation for practical autonomous navigation. The code has been released at \url{https://github.com/LuoXubo/StepNav}.
\end{abstract}

\section{Introduction}
Autonomous robots navigating complex, unstructured environments, such as cluttered forests or urban streets, must generate smooth, feasible trajectories from visual inputs like a history of observations and a goal image~\cite{wigness2019rugd, liang2022accurate}. These trajectories need to respect physical constraints (e.g., continuity and collision avoidance) while accounting for perceptual uncertainty arising from occlusions, visual ambiguities, or multi-modal routing choices (e.g., left vs. right at a junction). Classical approaches often decouple perception and planning, leading to brittle pipelines~\cite{tang2022perception}. In contrast, end-to-end generative models, such as diffusion policies~\cite{sridhar2024nomad} and conditional flow matching approaches~\cite{gode2025flownav}, have shown promise by directly synthesizing trajectories from visual data. However, these models are hampered by a fundamental challenge: they fail to produce physically plausible, uncertainty-aware trajectories efficiently, especially under visual ambiguity~\cite{janny2025reasoning}.

\begin{figure}[htp]
\centering
\includegraphics[width=\linewidth]{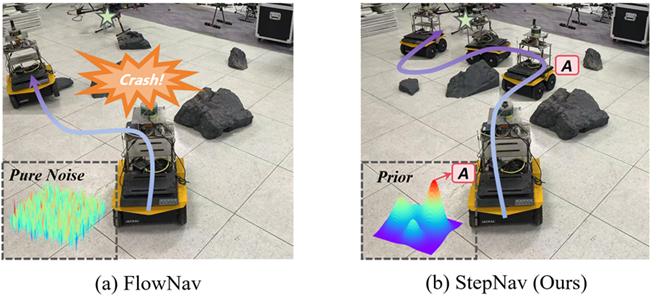}
\caption{(a) FlowNav failed to avoid obstacles when navigating towards the goal (an UAV). (b) Our StepNav estimates the success probability field to generate the prior trajectory that can accelerate the refinement and improve the success rate.}
\label{fig:intro}
\end{figure}

Unlike traditional motion planning problems where the state space and constraints can be explicitly modeled, visual navigation faces an inherent gap between high-dimensional, noisy perceptual inputs and the low-dimensional manifold of feasible trajectories. Bridging this gap requires mathematical structure: the model must simultaneously (i) extract temporally consistent dynamics from raw video, (ii) represent the multimodal nature of future possibilities under perceptual ambiguity, and (iii) generate smooth, safe trajectories fast enough for closed-loop execution. These three requirements often conflict: methods that emphasize diversity lose efficiency, those that enforce continuity ignore uncertainty, and those that are real-time often compromise on safety.

This challenge manifests in two intertwined limitations. First, trajectory generation typically starts from an unstructured Gaussian noise prior, which is agnostic to the manifold of valid trajectories. This disconnect necessitates a long refinement process (e.g., more than 10 diffusion steps) to sculpt noise into a coherent path, rendering them unsuitable for real-time robotics~\cite{ren2025prior, song2023consistency}. Second, such priors overlook the geometric structure inherent in visual inputs, such as ambiguous corridors or occluded obstacles, resulting in overconfident trajectories that ignore alternative paths and risk safety in dynamic environments. While learned priors have been explored~\cite{ren2025prior}, they often require auxiliary training stages and lack explicit mechanisms for capturing multi-modal path choices directly from visual cues. 

In summary, the core difficulty lies in constructing trajectory generators that are \emph{dynamically consistent, uncertainty-aware, and real-time feasible}. Existing approaches, by relying on unstructured priors, struggle with the fundamental trade-off between generative diversity and structured, efficient planning. This motivates our work: a principled method that injects structured, multimodal knowledge of navigation feasibility directly into the generative process, avoiding the inefficiency and brittleness of noise-driven sampling.

To tackle this challenge, we propose \textbf{StepNav}, a visual navigation framework that unifies variational field estimation and optimal-control-based generative refinement. Our key insight is that a dense, learned representation of navigability can serve as a powerful foundation for constructing structured priors. Our contributions are threefold:
\begin{itemize}
    \item We introduce a geometry-aware success probability field, learned via a biharmonic-regularized PDE, that captures the topology of all feasible navigation corridors from temporally-smoothed visual features.
    \item We propose a method to extract a structured, multi-modal mixture of trajectory candidates directly from this field by identifying low-energy paths, providing a powerful initialization for generative refinement.
    \item We formulate the final refinement as a regularized conditional flow-matching (Reg-CFM) problem, which explicitly optimizes for trajectory smoothness and safety, leading to higher-quality plans in fewer integration steps.
\end{itemize}

\section{Related Work}
\label{sec:related}
\subsection{End-to-End Visual Navigation}
Visual navigation has evolved from classical modular pipelines, which separate mapping, localization, and planning, towards end-to-end learning-based approaches. Early learning methods often relied on reinforcement learning (RL)~\cite{zeng2020survey,kulhanek2021visual}, but could struggle with sample efficiency and generalization. More recent approaches leverage large-scale pretraining and sophisticated architectures to improve robustness. For instance, ViNT~\cite{shah2023vint} introduced a foundation model for visual navigation by pre-training on diverse datasets. Other works have focused on incorporating explicit 3D representations, such as Gaussian Splatting, to provide better spatial grounding for the navigation policy~\cite{guo2025igl,lei2025gaussnav}. While these methods have advanced the state of the art, they often produce deterministic plans, limiting their ability to handle inherent environmental ambiguity.

\subsection{Generative Models for Trajectory Planning}
To address the limitations of deterministic planners, generative models, particularly score-based diffusion and flow-matching models, have emerged as a powerful paradigm for trajectory generation~\cite{carvalho2023motion,ke20243d}. These models can capture complex, multi-modal distributions of feasible paths. NoMaD~\cite{sridhar2024nomad} was a pioneering work that applied diffusion models to generate action sequences for navigation directly from visual inputs. However, a fundamental challenge in these models is their reliance on an uninformative prior, typically isotropic Gaussian noise. Starting from pure noise requires a long and computationally expensive reverse diffusion process to generate a structured trajectory, which can compromise physical plausibility and real-time performance.

Subsequent works have attempted to mitigate this issue by using more informative priors. NaviBridger~\cite{ren2025prior}, for example, uses a denoising diffusion bridge model initialized from a separately trained motion prior to shorten the generation process. NaviD~\cite{zhang2024navid} incorporates depth information to better constrain the generated paths. To obtain a faster inference speed, FlowNav~\cite{gode2025flownav} utilizes a combination of CFM and depth priors from Depth Anything-v2. However, these learned priors require additional training stages and do not explicitly model the geometric uncertainty that arises from the alignment between visual perception and physical space. A common limitation persists: the priors are either unstructured or lack a rigorous geometric foundation.

Our work directly addresses this critical gap. Instead of relying on unstructured noise or a separately learned network, we propose a unified, geometry-aware prior constructed on-the-fly from a learned success probability field computed over temporally refined video features. By extracting salient peaks and tracing low-energy corridors in this field we generate a compact, multi-modal mixture of candidate trajectories that serve as an informative initialization for a lightweight conditional flow-matching refinement. This design captures perceptual uncertainty, avoids costly auxiliary pretraining, and dramatically shortens refinement compared to noise-initialized diffusion methods, enabling efficient, safe, and diverse trajectory generation for real-time visual navigation. 

\begin{figure*}[htp]
\centering
\includegraphics[width=\linewidth]{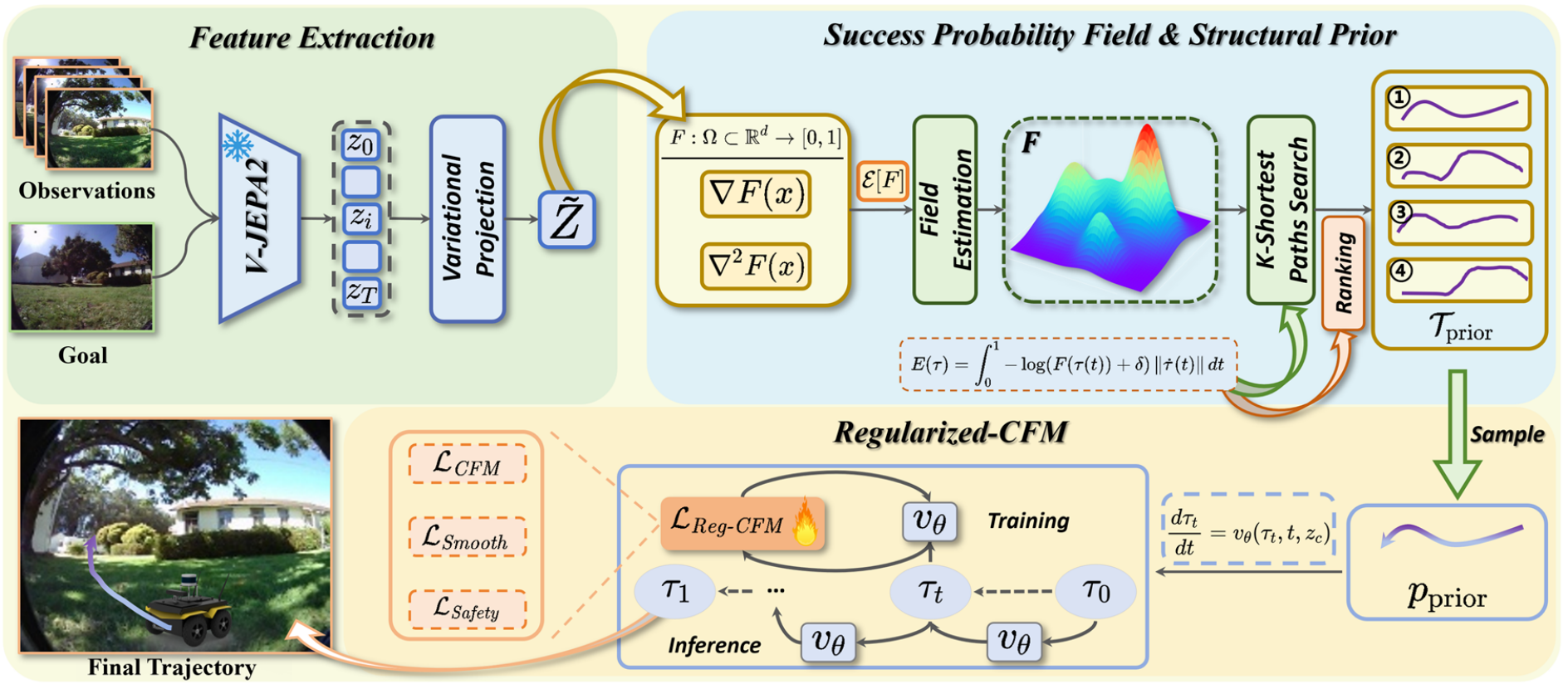}
\caption{Overview of StepNav. Given a sequence of input images and a goal image, we first extract temporal features via V-JEPA2 and refine them using our DIFP module. These refined features are used to predict a success probability field $F$. From this field, we estimate several prior trajectories and sample the prior trajectory according to the energy landscape $E(\tau)$. This prior is then refined into a smooth, feasible trajectory via Reg-CFM.}
\label{fig:pipeline}
\end{figure*}

\section{Method}
\label{sec:method}
We present \textbf{StepNav}, a three-stage framework that turns raw visual observations into an efficient, uncertainty-aware trajectory suitable for real-time navigation: (1) dynamics-inspired video feature refinement (DIFP); (2) unified success-field-driven structured prior synthesis; and (3) regularized conditional flow-matching (Reg-CFM) refinement. Formally, let the robot trajectory be a continuous curve
\begin{equation}
\tau: [0,1] \to \mathbb{R}^d,\quad \tau(0)=x_{\text{init}},\;\;\tau(1)=x_{\text{goal}}.
\end{equation}
We define a composite cost functional
\begin{equation}
    \begin{aligned}
        &J(\tau) =\\ 
        & \int_0^1 \Big( \lambda_g \,\ell_g(\tau(t)) + \lambda_s \,\ell_{\text{smooth}}(\dot \tau(t),\ddot \tau(t)) + \lambda_c \,\ell_{\text{coll}}(\tau(t)) \Big)\, dt,
    \end{aligned}
\end{equation}
where $\ell_g$ measures goal alignment, $\ell_{\text{smooth}}$ penalizes curvature and jerk, and $\ell_{\text{coll}}$ encodes collision risk. StepNav progressively constructs an approximation of the minimizer of $J(\tau)$ through increasingly structured stages.

\subsection{Feature Extraction and Refinement (DIFP)}
We first encode the input image history $\{I_t\}_{t=1}^T$ and the goal image $I_{\text{goal}}$ with a pre-trained V-JEPA2 encoder, yielding features 
\[
Z = [z_1,\dots,z_T] \in \mathbb{R}^{T \times D}.
\] 
To mitigate flicker and ensure temporal coherence, we formulate refinement as a projection problem:
\begin{equation}
\tilde Z = \arg\min_{Z'} \;\Big( \|Z'-Z\|_F^2 + \|LZ'\|_F^2 + \|Z' - z_c \mathbf{1}^\top\|_F^2 \Big),
\end{equation}
where $L$ is the temporal Laplacian enforcing smoothness, and $z_c=\sum_{t=1}^T w_t z_t$ is a global motion context with $w_t \propto \exp(v_t)$.  

The Euler-Lagrange condition gives the closed-form solution, which balances fidelity to the original features (first term), temporal smoothness (second term), and alignment with the global motion context (third term):
\begin{equation}
\tilde Z = (2I + L^\top L)^{-1}(Z + z_c \mathbf{1}^\top),
\end{equation}
ensuring the refined embeddings $\tilde z_t$ are both temporally smooth and goal-aware.

\subsection{Success Probability Field and Structured Prior}
From $(\tilde Z, z_c)$, a lightweight convolutional decoder head predicts a continuous success probability field $F:\Omega\subset \mathbb{R}^d \to [0,1]$ that encodes local navigability.  {This field serves as a learned, spatially-aware approximation of the inverse of the collision cost, $1-\ell_{\text{coll}}(x)$.} We define $F$ as the minimizer of the variational energy
\begin{equation}
    \begin{aligned}
    &\mathcal{E}[F] =\\ 
    &\int_\Omega \Big( (F(x)-y(x))^2 + \mu \|\nabla F(x)\|^2 + \nu \|\nabla^2 F(x)\|^2 \Big)\, dx,
    \end{aligned}
\end{equation}
where $y(x)$ are binary labels from expert demonstrations. The associated Euler-Lagrange equation yields a biharmonic-regularized Poisson PDE for $F$:
\begin{equation}
- \nu \Delta^2 F(x) + \mu \Delta F(x) + (F(x)-y(x))=0.
\end{equation}
Intuitively, the $\|\nabla F\|^2$ term enforces smoothness, while the biharmonic term $\|\nabla^2 F\|^2$ penalizes high curvature in the field, encouraging the formation of wide, smooth "corridors" of high success probability, which are ideal for path planning.

Given $F$, we define an energy landscape
\begin{equation}
E(\tau) = \int_0^1 -\log(F(\tau(t))+\delta)\, \|\dot \tau(t)\|\, dt,
\end{equation}
which measures the \textit{difficulty} of traversing a path $\tau$. Structured candidate priors are extracted as approximate minimizers of $E(\tau)$ under boundary conditions $\tau(0)=x_{\text{init}}, \tau(1)=x_{\text{goal}}$. Practically, this is solved by discretizing $\Omega$ into a graph and applying a K-shortest path algorithm.

To preserve multi-modality, we select a diverse subset $\mathcal{T}_{\text{prior}}=\{\tau^{(m)}\}_{m=1}^M$ using a greedy max-min Hausdorff criterion. This yields a mixture prior distribution:
\begin{equation}
p_{\text{prior}}(\tau) = \sum_{m=1}^M \pi_m\, \delta(\tau - \tau^{(m)}), \quad 
\pi_m \propto \exp\!\left(\tfrac{S(\tau^{(m)})}{T}\right),
\end{equation}
where $S(\tau^{(m)})$ scores success probability, path length, and curvature.

\subsection{Regularized Conditional Flow-Matching (Reg-CFM)}

While the structured prior provides feasible initial trajectories, it does not fully guarantee smoothness or safety. 
We therefore refine samples using a conditional flow-matching model that is regularized to respect physical constraints.

Given an initial trajectory $\tau_0 \sim p_{\text{prior}}$, refinement is posed as an optimal control problem:
\begin{equation}
\frac{d\tau_t}{dt} = v_\theta(\tau_t,t,z_c), \quad t \in [0,1],
\end{equation}
where $v_\theta$ is the learned flow field and $z_c$ provides global context.

Standard flow-matching aligns $v_\theta$ with the target flow $u_t$ from a stochastic interpolant. 
To promote feasible navigation, we augment this loss with smoothness and safety regularizers:
\begin{equation}
\begin{split}
\min_{v_\theta} \; \mathbb{E}_{\tau_0,\tau_1,t}\Big[ 
& \| v_\theta(\tau_t,t,z_c) - u_t(\tau \mid \tau_0,\tau_1)\|^2 \\
& + \rho \|\dddot \tau_t\|^2 
+ \kappa \phi(\tau_t) \Big],
\end{split}
\end{equation}
where the second term penalizes high jerk and 
$\phi(\tau_t)=-\log(d(\tau_t)-\epsilon)$ is a log-barrier that prevents collisions by discouraging states close to obstacles.
 
For implementation, we use two explicit terms on discretized waypoints $\{x_k\}$:  
\begin{equation}
\mathcal{L}_{\text{smooth}} = 
\sum_{k=1}^{K-3}\|x_{k+3}-3x_{k+2}+3x_{k+1}-x_k\|_2^2,
\end{equation}
penalizing jerk via finite differences, and
\begin{equation}
\mathcal{L}_{\text{safe}} = - \sum_{k=1}^K \log(\max(d(x_k)-\epsilon,0)),
\end{equation}
which acts as a barrier against entering unsafe regions, where the distance to obstacles $d(x_k)$ is estimated from the success field as $d(x_k) \propto 1 - F(x_k)$.

The overall training loss is therefore
\begin{equation}
\mathcal{L}_{\text{Reg-CFM}} 
= \mathcal{L}_{\text{FM}} 
+ \rho \mathcal{L}_{\text{smooth}} 
+ \kappa \mathcal{L}_{\text{safe}}.
\end{equation}

At inference time, refinement reduces to integrating the learned flow:
\begin{equation}
\tau_{t+\Delta t} = \tau_t + \Delta t \, v_\theta(\tau_t,t,z_c),
\qquad \Delta t = \tfrac{1}{N},
\end{equation}
with initialization $\tau_0 \sim p_{\text{prior}}$. 
Since priors are already structured and multimodal, only a small number of integration steps $N$ is required, enabling real-time trajectory generation.

\section{Experiments}
\label{sec:experiments}
We conduct a comprehensive set of experiments to validate the performance of StepNav in a range of challenging and diverse environments. Our evaluation is designed to rigorously answer three central questions corresponding to our core contributions:\\
\textbf{Overall Efficacy:} Does StepNav outperform other SOTA visual navigation methods on challenging, standard benchmarks in terms of success, safety, and path quality?\\
\textbf{Component Necessity:} Are the key components of our framework (i.e., the DIFP feature refinement, the structured multi-modal prior, and the Reg-CFM) all critical to its performance?\\
\textbf{Real-Time Feasibility:} Is the proposed method efficient enough for potential real-time deployment on robotic hardware, and how does its latency compare to other generative models?

\subsection{Experimental Setup}

\noindent\textbf{Datasets.} Following~\cite{sridhar2024nomad}, we aggregate the training datasets including RECON \cite{shah2021rapid}, SCAND \cite{karnan2022socially}, GoStanford \cite{hirose2019deep}, and SACSoN \cite{hirose2023sacson}. These datasets encompass over 1,450 total scenes, covering varied challenges like occlusions, dynamic obstacles, and lighting variations.

To ensure a robust evaluation, our primary quantitative comparisons are performed on two standard and challenging benchmarks that represent distinct domains: \textit{Indoor (Stanford 2D-3D-S)~\cite{armeni2017joint}} and \textit{Outdoor (Gazebo citysim)~\cite{koenig2004design}}. This collection presents challenges such as severe occlusions, dynamic obstacles, ambiguous corridors, and varied lighting. 

\begin{figure*}[htp]
\centering
\includegraphics[width=\linewidth]{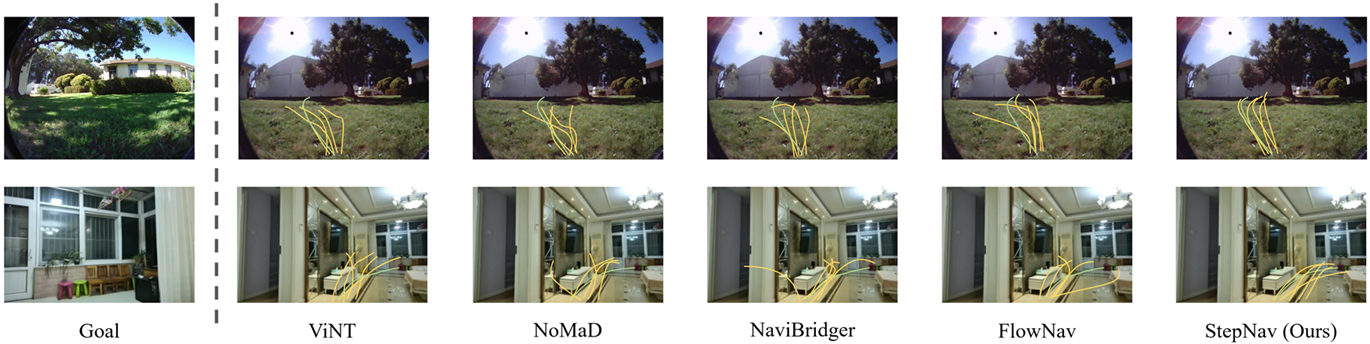}
\caption{Examples of StepNav in complex indoor and outdoor navigation tasks. The top row shows outdoor navigation in a simulated urban environment, where StepNav successfully navigates a multi-lane intersection with other agents. The bottom row illustrates indoor navigation in a cluttered room, demonstrating avoidance of static obstacles. The \textcolor{yellow}{yellow} trajectories are the estimated paths and the \textcolor{green}{green} dots represent the groundtruth waypoints.}
\label{fig:visualization}
\end{figure*}

\noindent\textbf{Baselines.} We compare StepNav against a representative set of SOTA methods, including both generative and deterministic approaches: \textit{ViNT}~\cite{shah2023vint}: A strong deterministic visual navigation model based on a pretrained foundation model. \textit{NoMaD}~\cite{sridhar2024nomad}: A pioneering diffusion-based policy for navigation that starts from an unstructured Gaussian noise prior. \textit{NaviBridger}~\cite{ren2025prior}: An improved diffusion model that uses a separately trained motion prior to initialize a denoising diffusion bridge, aiming for higher efficiency. \textit{FlowNav}~\cite{gode2025flownav}: A conditional flow matching model that refines trajectories from Gaussian noise, without structured priors or regularization.

For fair comparison, all learning-based methods are trained on our specific dataset splits using the authors' official codebases and recommended hyperparameters.

\noindent\textbf{Tasks and Protocols.} We evaluate all methods on two distinct tasks, following the protocol established in~\cite{sridhar2024nomad}: \textit{Basic Task:} Standard point-goal navigation where agents are spawned in seen environments (from the training distribution of scenes) and must navigate to a specified goal.\textit{Adaptation Task:} A more challenging zero-shot transfer task where agents must navigate in entirely unseen environments with significant domain shifts (e.g., different lighting, weather conditions, and object textures) to test generalization. An episode is considered successful if the agent reaches within 0.2 meters of the goal in under 500 steps.

\noindent\textbf{Metrics.} We report four standard metrics to comprehensively assess the performance:
\begin{itemize}
    \item \textit{Success Rate (SR, \% $\uparrow$):} The percentage of episodes where the agent reaches the goal within a predefined tolerance. The primary measure of task completion.
    \item \textit{Success Weighted by Path Length (SPL, $\uparrow$):} A standard navigation metric~\cite{anderson2018evaluation}, defined as $\mathrm{SPL}_i = S_i \cdot L_i / \max(P_i, L_i)$, where $S_i=1$ if the goal is reached (within 0.2\,m, $\leq$500 steps), else $0$; $L_i$ is the shortest path length, $P_i$ the executed path length. Reported SPL is the mean over episodes.
    \item \textit{Collision Rate (\%, $\downarrow$):} The percentage of episodes ending in a collision, directly measuring safety.
    \item \textit{Minimum Snap (MS, m$^2$/s$^7$ $\downarrow$):} A measure of the integral of the squared snap (fourth derivative of position), which quantifies trajectory smoothness. Lower values indicate more dynamically feasible and comfortable paths.
\end{itemize}

\noindent\textbf{Implementation Details.} All models were trained on NVIDIA RTX 4090. We will release our code and evaluation framework upon publication to ensure full reproducibility. We deployed StepNav on a real-world Clearpath Jackal robot equipped with an NVIDIA Jetson AGX Orin, demonstrating real-time navigation in both indoor and outdoor settings. The robot uses a forward-facing RGB camera for visual input and relies on onboard computation for trajectory generation and execution.

\begin{table*}[htp]
\centering
\caption{Quantitative results reported as mean $\pm$ standard deviation on indoor (Stanford 2D-3D-S) and outdoor (Citysim) navigation tasks.}
\label{tab:main_results}
\begin{tabular}{c l c c c c c c c c}
\toprule
\multirow{2}{*}{\textbf{Task}} & 
\multirow{2}{*}{\textbf{Method}} & 
\multicolumn{4}{c}{\textbf{Indoor (Stanford 2D-3D-S)}} & 
\multicolumn{4}{c}{\textbf{Outdoor (Citysim)}} \\
\cmidrule(lr){3-6} \cmidrule(lr){7-10}
& & \textbf{SR (\%) $\uparrow$} & \textbf{SPL $\uparrow$} & \textbf{Coll. (\%) $\downarrow$} & \textbf{MS (m$^2$/s$^7$) $\downarrow$} & \textbf{SR (\%) $\uparrow$} & \textbf{SPL $\uparrow$} & \textbf{Coll. (\%) $\downarrow$} & \textbf{MS (m$^2$/s$^7$) $\downarrow$} \\
\midrule
\multirow{5}{*}{\makecell{Basic\\Task}} & 
ViNT & 68$\pm$2.7 & 0.71$\pm$0.04 & 1.0$\pm$0.25 & 0.28$\pm$0.04 & 20$\pm$3.1 & 0.68$\pm$0.05 & 0.8$\pm$0.22 & 0.35$\pm$0.05 \\
& NoMaD & 86$\pm$2.5 & 0.69$\pm$0.04 & 0.7$\pm$0.24 & 0.30$\pm$0.04 & 22$\pm$3.4 & 0.65$\pm$0.05 & 0.6$\pm$0.21 & 0.37$\pm$0.05 \\
& NaviBridger & 92$\pm$1.9 & 0.73$\pm$0.03 & 0.6$\pm$0.18 & 0.26$\pm$0.03 & 44$\pm$3.6 & 0.69$\pm$0.04 & 0.6$\pm$0.20 & 0.33$\pm$0.04 \\
& FlowNav & 90$\pm$2.2 & \textbf{0.81$\pm$0.02} & 0.8$\pm$0.22 & 0.22$\pm$0.03 & 40$\pm$3.1 & 0.73$\pm$0.04 & \textbf{0.5$\pm$0.18} & 0.30$\pm$0.04 \\
& \cellcolor{mygreen}StepNav & \cellcolor{mygreen}\textbf{95$\pm$1.1} & \cellcolor{mygreen}{0.80$\pm$0.02} & \cellcolor{mygreen}\textbf{0.6$\pm$0.12} & \cellcolor{mygreen}\textbf{0.20$\pm$0.02} & \cellcolor{mygreen}\textbf{57$\pm$3.0} & \cellcolor{mygreen}\textbf{0.76$\pm$0.03} & \cellcolor{mygreen}\textbf{0.5$\pm$0.10} & \cellcolor{mygreen}\textbf{0.28$\pm$0.03} \\
\midrule
\multirow{5}{*}{\makecell{Adaptation\\Task}} & 
ViNT & 28$\pm$3.4 & 0.63$\pm$0.05 & 1.6$\pm$0.35 & 0.33$\pm$0.04 & 38$\pm$3.8 & 0.60$\pm$0.06 & 0.4$\pm$0.13 & 0.40$\pm$0.05 \\
& NoMaD & 32$\pm$3.7 & 0.61$\pm$0.05 & 1.3$\pm$0.33 & 0.35$\pm$0.04 & 52$\pm$4.1 & 0.58$\pm$0.06 & 0.3$\pm$0.12 & 0.42$\pm$0.05 \\
& NaviBridger & 88$\pm$2.2 & 0.65$\pm$0.04 & 0.5$\pm$0.20 & 0.31$\pm$0.03 & 64$\pm$4.2 & 0.62$\pm$0.05 & 0.3$\pm$0.11 & 0.38$\pm$0.04 \\
& FlowNav & 85$\pm$2.5 & 0.71$\pm$0.04 & 0.6$\pm$0.21 & 0.33$\pm$0.03 & 65$\pm$4.0 & 0.70$\pm$0.05 & 0.4$\pm$0.12 & 0.36$\pm$0.04 \\
& \cellcolor{mygreen}StepNav & \cellcolor{mygreen}\textbf{90$\pm$1.6} & \cellcolor{mygreen}\textbf{0.74$\pm$0.03} & \cellcolor{mygreen}\textbf{0.4$\pm$0.12} & \cellcolor{mygreen}\textbf{0.24$\pm$0.02} & \cellcolor{mygreen}\textbf{68$\pm$3.5} & \cellcolor{mygreen}\textbf{0.71$\pm$0.04} & \cellcolor{mygreen}\textbf{0.3$\pm$0.10} & \cellcolor{mygreen}\textbf{0.32$\pm$0.03} \\
\bottomrule
\end{tabular}
\end{table*}

\subsection{Comparison with SOTA Methods}
As summarized in Table~\ref{tab:main_results}, StepNav demonstrates a clear performance advantage over all baselines across both indoor and outdoor navigation tasks. On the indoor \textit{Basic Task}, StepNav achieves an SR of 95\%, outperforming the next best baseline (NaviBridger) by 3 percentage points. This performance gain is even more pronounced in the more challenging Adaptation Task. Crucially, the improvement is not just in the success rate. StepNav also achieves the highest SPL scores, indicating that its successful paths are more efficient. Furthermore, it records the lowest Minimum Snap and is tied for the lowest Collision Rate, demonstrating its ability to generate trajectories that are both safer and smoother. This superior performance stems from the structured prior, which provides a strong, geometry-aware initialization. Unlike NoMaD, which must expend many refinement steps to shape unstructured noise, or NaviBridger, whose separately trained prior is inherently unimodal, StepNav's on-the-fly, multi-modal prior allows it to robustly handle ambiguous scenarios (e.g., junctions), leading to more effective and safer plans.

\begin{figure}[htp]
    \centering
    \includegraphics[width=0.95\linewidth]{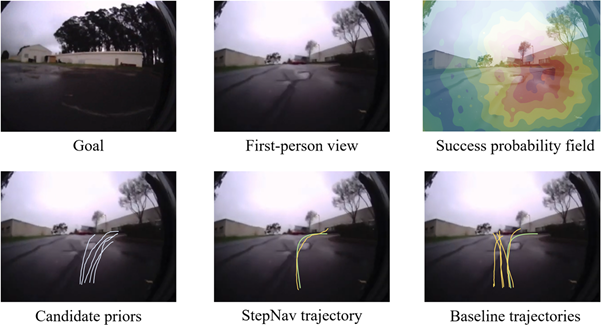}
    \caption{A qualitative visualization of StepNav's core mechanism at an ambiguous T-junction. The gray lines indicate the generated prior trajectories, while the yellow lines represent the final trajectories of StepNav and other baseline models. The ground truth trajectories are shown in green.}
    \label{fig:qualitative_multimodal}
\end{figure}

The efficacy of our multi-modal prior is particularly evident in ambiguous scenarios. Figure~\ref{fig:qualitative_multimodal} provides a qualitative visualization of StepNav's internal mechanism when encountering a challenging T-junction. At this junction, the learned success probability field correctly identifies two viable corridors: one straight ahead and another to the right. Subsequently, our prior extraction method generates distinct candidate trajectories for each potential corridor (gray lines). The highest-scoring candidate, based on our energy function, is then selected and refined via Reg-CFM into the smooth, final trajectory (yellow line). In contrast, baseline methods that depend on unimodal or unstructured priors often commit prematurely to a single, and potentially incorrect, path in such situations.

\subsection{Ablation Studies}
To deconstruct StepNav's performance and validate our design choices, we conduct a series of ablation experiments on the Stanford 2D-3D-S dataset, chosen for its diverse set of indoor challenges.

\noindent\textbf{1. Impact of the Structured Prior.} This is the most critical ablation, designed to test the value of our success-field-driven prior. We compare our full model against two variants:
\begin{itemize}
    \item \textit{Gaussian Prior:} Our Reg-CFM initialized with a standard isotropic Gaussian noise, effectively making it similar to a flow-based version of NoMaD.
    \item \textit{Peaks as Prior:} A simplified, unimodal prior formed by only using the detected peaks from the success probability field and connecting them.
\end{itemize}
Results in Table~\ref{tab:ablation_prior} show a clear and significant performance hierarchy. The \textit{Gaussian Prior} variant performs worst, confirming that an uninformative prior struggles to generate successful and efficient trajectories. Using even a simple structured prior (\textit{Peaks as Prior}) yields a massive improvement (e.g., +12.8\% SR), demonstrating the power of grounding the initial trajectory in the success field. Finally, our full multi-modal prior provides another substantial boost, particularly in SR and SPL, which validates the importance of exploring diverse hypotheses via low-energy corridor tracing.

\begin{table}[htp]
\centering
\caption{Ablation study on the components of the structured prior.}
\label{tab:ablation_prior}
\begin{tabular}{lcccc}
\toprule
\textbf{Prior Type} & \textbf{SR (\%) $\uparrow$} & \textbf{SPL $\uparrow$} & \textbf{Coll. (\%) $\downarrow$} & \textbf{MS $\downarrow$} \\
\midrule
Gaussian Prior & 72.5 & 0.64 & 1.3 & 0.38 \\
Peaks as Prior & 85.3 & 0.71 & 0.8 & 0.31 \\
\rowcolor{mygreen}
\textbf{StepNav (Full)} & \textbf{94.8} & \textbf{0.80} & \textbf{0.6} & \textbf{0.20} \\
\bottomrule
\end{tabular}
\end{table}

\noindent\textbf{2. Impact of Feature Refinement.} To verify the contribution of the DIFP module, we compare our full model with a variant, \textit{w/o DIFP}, which uses the raw V-JEPA2 features directly for success field prediction. As shown in Table~\ref{tab:ablation_feature}, removing DIFP causes a stark degradation across all metrics, with SR dropping by 15\%. This confirms our hypothesis that raw video features contain temporal inconsistencies and noise that are detrimental to planning. The DIFP module's ability to enforce motion-coherence is therefore crucial for downstream performance.

\begin{table}[htp]
\centering
\caption{Ablation study on the DIFP feature refinement module.}
\label{tab:ablation_feature}
\begin{tabular}{lcccc}
\toprule
\textbf{Variant} & \textbf{SR (\%) $\uparrow$} & \textbf{SPL $\uparrow$} & \textbf{Coll. (\%) $\downarrow$} & \textbf{MS $\downarrow$} \\
\midrule
w/o DIFP & 79.8 & 0.61 & 1.9 & 0.29 \\
\rowcolor{mygreen}
\textbf{StepNav (Full)} & \textbf{94.8} & \textbf{0.80} & \textbf{0.6} & \textbf{0.20} \\
\bottomrule
\end{tabular}
\end{table}

\noindent\textbf{3. Ablation on the Success Probability Field.} To demonstrate that our learned, dense success probability field is a superior representation for prior generation, we compare it against simpler alternatives, keeping all other framework components (DIFP, path extraction logic, CFM optimizer) fixed.
\begin{itemize}
    \item \textit{Direct Waypoint Regression:} This variant removes the concept of a field entirely. The model head is modified to directly regress a sequence of key waypoints, testing if a dense intermediate representation is necessary.
    \item \textit{Depth Field:} We use a pre-trained monocular depth estimator (DepthAnything) to create a simple geometric traversability field where higher values correspond to more distant (i.e., open) space. This tests if simple geometric cues are sufficient, as opposed to a learned, task-oriented field.
    \item \textit{StepNav (Learned Field):} Our full proposed method.
\end{itemize}
Table~\ref{tab:ablation_field} shows that our learned field significantly outperforms both alternatives. Direct regression performs poorly, as it lacks the rich topological context of a dense field, making it brittle. The depth field performs better but is still inferior to our learned field. This is because our field is not just aware of "empty space" but is trained end-to-end to identify regions that are part of a "successful path to the goal," embedding crucial task-oriented semantics that pure geometry lacks. This experiment confirms that learning a dense, task-aware success field is a key innovation of our approach. Notably, the depth field achieves slightly better performance in terms of collision rate, suggesting that explicit depth cues may provide complementary benefits. This insight motivates future work on integrating depth information more directly into our framework.

\begin{table}[htp]
\centering
\caption{Ablation study on the representation used for prior generation.}
\label{tab:ablation_field}
\begin{tabular}{lccc}
\toprule
\textbf{Prior Generation Method} & \textbf{SR (\%) $\uparrow$} & \textbf{SPL $\uparrow$} & \textbf{Coll. (\%) $\downarrow$} \\
\midrule
Waypoint Regression & 74.2 & 0.66 & 1.6 \\
Depth Field & 79.5 & 0.73 & 0.5 \\
\rowcolor{mygreen}
\textbf{StepNav} & \textbf{94.8} & \textbf{0.80} & \textbf{0.6} \\
\bottomrule
\end{tabular}
\end{table}

\noindent\textbf{4. Impact of Refinement Strategy.} Finally, we validate our choice of the Reg-CFM for refinement. Using our complete structured prior as a fixed starting point, we compare: 
\begin{itemize}
    \item Vanilla \textit{DDIM}~\cite{song2020denoising} (standard diffusion-based refinement).
    \item Vanilla \textit{CFM}~\cite{lipman2022flow} (no extra regularizers).
    \item \textit{CFM+$\mathcal{L}_{\text{smooth}}$} (only smoothness penalty).
    \item \textit{CFM+$\mathcal{L}_{\text{safe}}$} (only safety barrier).
    \item \textit{Reg-CFM} (both $\mathcal{L}_{\text{smooth}}$ \& $\mathcal{L}_{\text{safe}}$).
\end{itemize}
As shown in Table~\ref{tab:ablation_refinement}, compared to vanilla CFM, incorporating only the $\mathcal{L}_{\text{smooth}}$ term improves all metrics, most notably reducing the Minimum Snap (MS from 0.30 to 0.23) and more than halving the collision rate (from 2.1\% to 0.8\%). Conversely, adding only the $\mathcal{L}_{\text{safe}}$ term significantly lowers the collision rate to just 0.6\%, but it has no effect on smoothness. When both regularizers are used together, our method achieves the best overall performance, with further improvements in SR and SPL, and the lowest Collision rate and MS simultaneously. This demonstrates that the two constraints are complementary. Even with a strong initial prior, explicit and task-relevant regularization yields improvements in deployable quality.

\begin{table}[htp]
\centering
\caption{Refinement strategy ablation (all initialized by the same structured prior; 10 refinement steps unless otherwise stated).}
\label{tab:ablation_refinement}
\begin{tabular}{lcccc}
\toprule
\textbf{Strategy} & \textbf{SR (\%) $\uparrow$} & \textbf{SPL $\uparrow$} & \textbf{Coll. (\%) $\downarrow$} & \textbf{MS $\downarrow$} \\
\midrule
DDIM (10 steps)            & 84.1 & 0.68 & 1.9 & 0.33 \\
CFM (10 steps)             & 84.9 & 0.71 & 2.1 & 0.30 \\
CFM+$\mathcal{L}_{\text{smooth}}$ & 89.0 & 0.77 & 0.8 & 0.23 \\
CFM+$\mathcal{L}_{\text{safe}}$   & 87.1 & 0.75 & 0.6 & 0.30 \\
\rowcolor{mygreen}
\textbf{Reg-CFM (Ours)}    & \textbf{94.8} & \textbf{0.80} & \textbf{0.6} & \textbf{0.20} \\
\bottomrule
\end{tabular}
\end{table}

\begin{figure}[htp]
    \centering
    \includegraphics[width=0.95\linewidth]{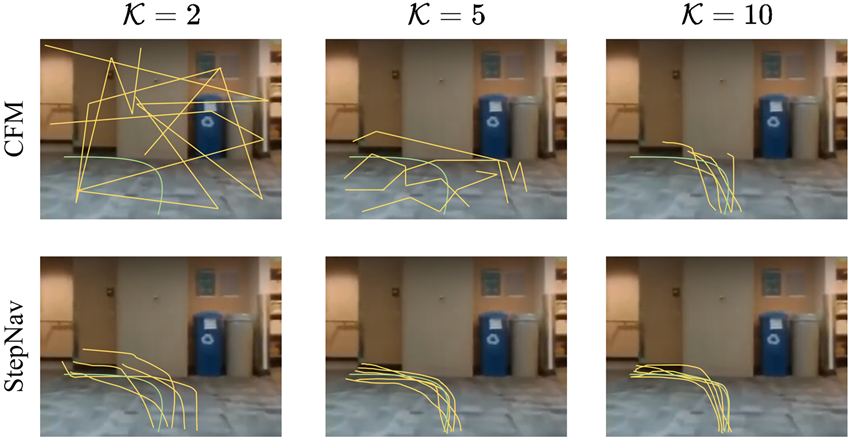}
    \caption{Refinement strategies under identical structured prior initialization. StepNav converges faster and yields smoother trajectories compared to CFM, which starts from Gaussian noise.}
    \label{fig:ablation_flow}
\end{figure}

We also visualize the evolution of the trajectories across 2, 5, and 10 refinement steps. We compare StepNav with the standard CFM baseline, which uses the same architecture but is initialized from Gaussian noise. As shown in Fig.~\ref{fig:ablation_flow}, StepNav's structured prior already captures the global topology of the scene, allowing it to quickly refine to a high-quality trajectory within just a few steps. In contrast, CFM starts from an unstructured prior and requires many more steps to converge to a reasonable path. Even after 10 steps, CFM's trajectory remains suboptimal and less smooth compared to StepNav. This visualization confirms that our structured prior significantly accelerates convergence and leads to superior final trajectories.

\begin{figure}[htp]
    \centering
    \includegraphics[width=\linewidth]{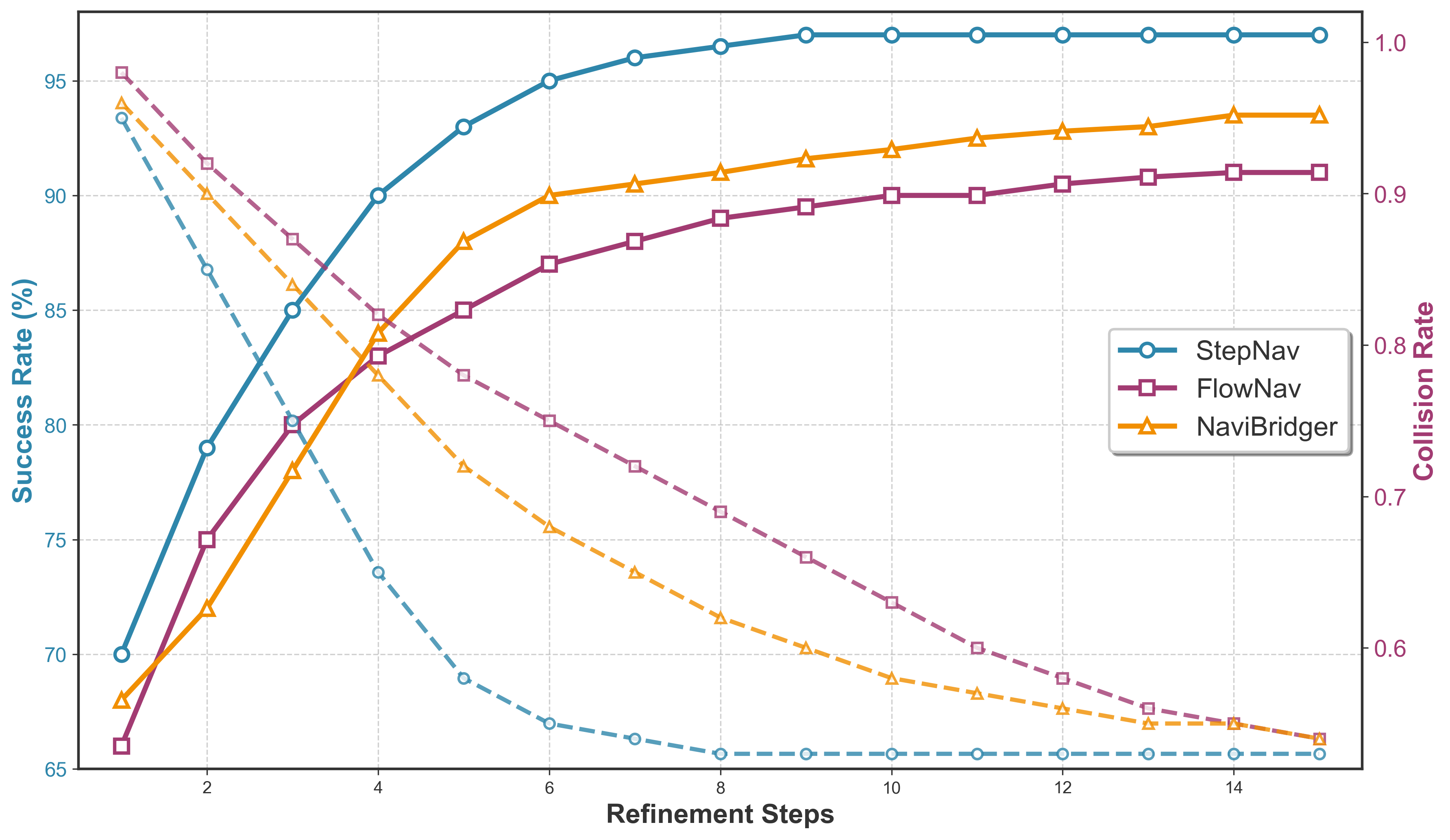}
    \caption{Success rate (SR, dashed lines) and collision rate (Coll., solid lines) as functions of refinement steps. StepNav converges within five steps, whereas FlowNav and NaviBridger require more iterations to approach comparable performance.}
    \label{fig:steps_scaling}
\end{figure}

\noindent\textbf{5. Ablation on Refinement Steps.}
We evaluate the effect of refinement steps on scalability. As shown in Figure~\ref{fig:steps_scaling}, StepNav achieves rapid gains and converges within five iterations, reaching over 95\% SR with negligible collision rates. By contrast, FlowNav and NaviBridger require more than ten steps to attain comparable SR, while consistently exhibiting higher collision rates. These results indicate that StepNav yields faster convergence and more reliable navigation with fewer refinements.

\subsection{Real-World Deployment}
\begin{figure}[htp]
    \centering
    \includegraphics[width=\linewidth]{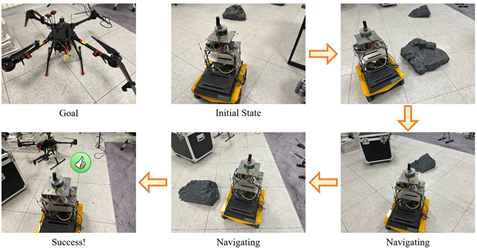}
    \caption{Real-world deployment of StepNav on a Clearpath Jackal robot navigating through an indoor environment.}
    \label{fig:real_robot}
\end{figure}

We deployed StepNav on a Clearpath Jackal robot equipped with an NVIDIA Jetson AGX Orin, demonstrating real-time navigation in complex indoor settings. The robot uses a forward-facing RGB camera for visual input and relies on Orin AGX for trajectory generation and execution. The system operates at 8.5Hz, surpassing FlowNav's 6.9Hz, thereby confirming its suitability for real-time applications. Fig.~\ref{fig:real_robot} shows the robot successfully navigating through cluttered environments, avoiding obstacles, and reaching specified goals. Real-world tests confirm that StepNav's efficient inference and robust planning capabilities effectively translate from simulation to physical deployment. 

\section{Conclusion}
\label{sec:conclusion}
This paper presents StepNav, a novel framework for visual navigation that combines a learned success probability field, a structured multi-modal prior, and a regularized conditional flow matching model. Our approach addresses key limitations of prior generative navigation methods by providing a strong, geometry-aware initialization and incorporating task-relevant regularization. Extensive experiments on challenging indoor and outdoor benchmarks demonstrate that StepNav significantly outperforms state-of-the-art baselines in terms of success rate, path efficiency, safety, and smoothness. Ablation studies confirm the necessity of each core component, and real-world deployment on a robotic platform validates its practical feasibility. Future work will focus on improving robustness in dynamic environments by integrating short-term motion forecasts directly into the success probability field and exploring semi-supervised or self-supervised methods to reduce the reliance on expert demonstrations.

\end{document}